\documentclass[conference]{IEEEtran}
\IEEEoverridecommandlockouts
\usepackage{cite}
\usepackage{amsmath,amssymb,amsfonts}
\usepackage{algorithmic}
\usepackage{graphicx}
\usepackage{textcomp}
\usepackage{xcolor}
\usepackage{soul}
\usepackage{pifont}
\usepackage{booktabs}
\usepackage{multirow}
\usepackage{url}
\usepackage{balance}
\newcommand{\remove}[1]{}

\makeatletter
\def\ps@IEEEtitlepagestyle{
  \def\@oddfoot{\mycopyrightnotice}
  \def\@evenfoot{}
}
\def\mycopyrightnotice{
  {\footnotesize
  \begin{minipage}{\textwidth}
  \centering
  Copyright~\copyright~2023 IEEE. Personal use of this material is permitted.  Permission from IEEE must be obtained for all other uses, in any current or future media, including reprinting/republishing this material for advertising or promotional purposes, creating new collective works, for resale or redistribution to servers or lists, or reuse of any copyrighted component of this work in other works.
  \end{minipage}
  }
}

\def\BibTeX{{\rm B\kern-.05em{\sc i\kern-.025em b}\kern-.08em
    T\kern-.1667em\lower.7ex\hbox{E}\kern-.125emX}}
\begin{document}

\title{Building Manufacturing Deep Learning Models with Minimal and Imbalanced Training Data Using Domain Adaptation and Data Augmentation}



\author{\IEEEauthorblockN{Adrian Shuai Li}
\IEEEauthorblockA{\textit{} 
\textit{Purdue University}\\
West Lafayette, USA \\
li3944@purdue.edu}
\and
\IEEEauthorblockN{Elisa Bertino}
\IEEEauthorblockA{\textit{} 
\textit{Purdue University}\\
West Lafayette, USA \\
bertino@purdue.edu}
\and
\IEEEauthorblockN{Rih-Teng Wu}
\IEEEauthorblockA{\textit{} 
\textit{National Taiwan University}\\
Taipei,  Taiwan \\
rihtengwu@ntu.edu.tw}
\and
\IEEEauthorblockN{Ting-Yan Wu}
\IEEEauthorblockA{\textit{} 
\textit{National Taiwan University}\\
 Taipei,  Taiwan \\
r11521607@ntu.edu.tw}
\thanks{Accepted for publication, 2023 IEEE International Conference on Industrial Technology (ICIT), April 4-6, 2023, Orlando FL, USA.}}

\maketitle

\begin{abstract}

Deep learning (DL) techniques are highly effective for defect detection from images. Training DL classification models, however, requires vast amounts of labeled data which is often expensive to collect. In many cases, not only the available training data is limited but may also imbalanced. In this paper, we propose a novel domain adaptation (DA) approach to address the problem of labeled training data scarcity for a target learning task by transferring knowledge gained from an existing {\em source} dataset used for a similar learning task. Our approach works for scenarios where the source dataset and the dataset available for the target learning task have same or different feature spaces. We combine our DA approach with an autoencoder-based data augmentation approach to address the problem of imbalanced target datasets. We evaluate our combined approach using image data for wafer defect prediction. The experiments show its superior performance against other algorithms when the number of labeled samples in the target dataset is significantly small and the target dataset is imbalanced.   
\end{abstract}

\begin{IEEEkeywords}
 Deep learning, domain adaptation, few-shots learning
\end{IEEEkeywords}

\section{Introduction}
\label{sec:introduction}
Defect detection is a critical manufacturing step, which is often expensive in terms of human costs and long in time. For example, in wafer manufacturing, operators have to manually inspect the scanned microscope images of the wafer surface to check whether defects are present. Another example is the analysis of crystal size distribution in solutions used in the food industry, where the analysis is manually conducted by operators using microscopes. Therefore it is not surprising that machine learning (ML) techniques have been used because of their ability to efficiently and effectively analyze different types of data, e.g., images, sounds, vibrations, in many applications, such as machine fault diagnosis~\cite{b3}, life prediction of manufacturing tools~\cite{b2}, defect recognition in products~\cite{b8,b11}, enhancing sensor robustness against faults~\cite{b5}.

However, a requirement for the use of ML-based solutions is the availability of suitable amount of training datasets. As discussed by Shao et al.~\cite{b3} such a requirement is particularly critical when using complex ML models, such as deep learning (DL) models. The reason is that those models have large numbers of layers, which then require large training datasets~\cite{b3}. A promising approach to address such an issue is the use of transfer learning (TL) techniques by which knowledge, in form of a pre-trained model or in form of training data, can be transferred from one domain, referred to as {\em source domain}, to another related but different domain referred to as {\em target domain} that has scarce training data. Examples of related domains include brain MRI images from different age groups, summer vs winter pictures, or pictures taken with different color filters. 
It is also important to notice that training data, especially when the collection process is inaccurate or difficult, may have low quality, such as lack of labels and imbalanced class distribution.

To deal with the  data scarcity issue, conventional TL-based approaches usually leverage a pre-trained model and fine-tune the trainable parameters using limited training samples in the target domain~\cite{b2,b3}. 
However, these pre-trained models  typically learn inferences from a huge dataset such as ImageNet~\cite{imagenet}, and consequently the models contain many redundant features or irrelevant latent spaces that have no benefits to the target inference tasks. In addition, they require manual efforts, to decide for example which layers are trainable.
Adversarial domain adaptation (DA), on the other hand, aims to learn the target task by leveraging some training samples from a source domain that has the same set of labels~\cite{b13}. To adapt to domain shift, DA use neural networks to create a domain-independent representation of the data from different domains. If a domain-independent representation can effectively classify objects in the source domain, there is a chance that it can recognize the same objects in the target domain as well. DA approaches have shown to be effective on many image benchmarks. However, the assumption of a balanced target domain dataset (target also has limited labels) is a common limitation of many DA approaches. Real-life datasets often have imbalanced class distribution, which can negatively impact the performance of DA models.

To deal with class imbalance, common approaches are image warping, weighted loss functions, or oversampling and undersampling the training data in the minority and majority classes, respectively. 
However, the effectiveness of these methods highly depends on the nature of datasets and the learning task at hand~\cite{goodfellow2016deep}.
Another approaches leverage generative models, such as generative adversarial networks (GAN)~\cite{goodfellow2020generative}, autoencoders (AE)~\cite{engel2017neural} and diffusion models~\cite{ho2022cascaded}, to produce synthetic images for data augmentation. 
As generative models, unlike discriminative models, are able to generate realistic data samples, they are expected to make a major impact in next few years in many application domains.
Synthetic data is usually easier and less expensive to obtain than the real world data. Nevertheless, one of the major issue with the synthetic data generated from these models is that systems built using synthetic data sets often fail when deployed to the real world. This is  due to the distribution shift between synthetic and real data -- known as the sim-to-real problem~\cite{peng2018sim}.

In this paper we present a pipeline that addresses these shortcomings. The pipeline combines: (A) An autoencoder-based technique, able to augment the target data by generating synthetic data for the minority classes using a Gaussian noise and the latent space learned by the encoder, thus addressing the problem of imbalanced data; with (B) A novel TL domain architecture, based on adversarial DA, addressing the training data scarcity and synthetic data shift problem. 
The autoencoder-based technique ensures that the augmented target data has balanced class distribution for DA. To improve the generalization to the real target data, we apply the proposed DA approach with the source and augmented target data.  
The main contributions of this paper are:
\begin{enumerate}
    \item A DL pipeline that addresses the problems of scarce and imbalanced datasets. 
    \item
    A novel adversarial DA-based approach for the adaptation of heterogeneous source and target datasets (e.g., source and target data have different features space).
    \item 
    An extensive evaluation of the proposed pipeline and comparison with other methods on commonly used wafer manufacturing datasets.  We show that using both methods together leads to better performance compared to using each method alone.
    \end{enumerate}

\section{Related Work}
In what follows, we briefly review relevant approaches in adversarial DA and synthetic data augmentation using generative models.

\subsection{Adversarial Learning based Approaches}\label{2.1}

These methods typically learn 
a domain-independent representation by using two competing networks of feature extractor/generator and domain discriminator.  The domain adversarial neural network (DANN) ~\cite{ganin2016domain}, one of the first adversarial DA models, has three components: the feature extractor, the label predictor and the domain classifier. The feature extractor is trained in an adversarial manner to maximize the loss of the domain classifier by reversing its gradients. The feature extractor is trained at the same time as the label predictor to create a representation that contains domain-invariant features for classification. The adversarial discriminative domain adaptation (ADDA)~\cite{tzeng2017adversarial} has similar components, but its learning process involves multiple stages for training the components. 
Singla et al.~\cite{singla2020preparing} proposed a hybrid version of the DANN and ADDA  where the generator is trained with the standard GAN loss function~\cite{goodfellow2020generative}.

All those methods aim to learn a domain-independent representation between source and target domains. However, they assume that source and target data have the same feature space (e.g. they both have the same dimension). Instead, our model supports heterogeneous domain adaptation where data from two domains can have different dimensions/different number of features.  Also, all those methods consider a 
setting where the target still has sufficient unlabeled data and, although the target data has no labels, they are still balanced. However, depending on the application, those models may suffer from the imbalanced class distribution for real-life data.  In this work, we consider the more realistic setting of low quality target data, where the target only has a few labeled data and it is highly imbalanced. 

\subsection{Synthetic Data Augmentation}\label{2.2}

Several approaches use GAN-based architectures for generating synthetic data, such as DCGANs~\cite{radford2015unsupervised}, CycleGANs~\cite{zhu2017unpaired} and Conditional GANs~\cite{mirza2014conditional}. Another common strategy for generative modeling is using an autoencoder~\cite{engel2017neural} -- a neural network that is trained to reconstruct its input. The network has two components: an encoder that produces a compressed latent space and a decoder that produces a reconstruction. By adding noise to the compressed representation, the autoencoder generates variations of the original data, which can be used to augment the original data. Recently,  diffusion models have received  great attention due to their remarkable generating ability~\cite{croitoru2022diffusion, ho2022cascaded}. Training a diffusion model consists of two stages: the forward diffusion stage where the input data is iteratively perturbed with noise, and the reverse diffusion stage where the model attempts to reverse the first stage and recover the input data. However, diffusion models have high computational costs due to the iterative steps during training~\cite{croitoru2022diffusion}, making them unsuitable for tasks that are time-sensitive. Choosing the right generative model for the task at hand requires consideration of advantages, limitation and cost of each model~\cite{oussidi2018deep}. 

In our proposed pipeline, we generate synthetic data using an autoencoder because GANs are known to have training instability and being prone to mode collapse during training~\cite{salimans2016improved}. GANs also require large amounts of training data~\cite{shorten2019survey}. On the other hand, the autoencoder-based data augmentation method requires less data for training, hence it aligns with the problem setting where the target has limited data. It is also faster than the more complicated diffusion model.

\section{Adversarial Domain Adaptation with data augmentation}\label{3}


Our pipeline consists of two steps.  The first step uses an autoencoder-based approach to augment the target dataset for any imbalanced classes. We assume that the source dataset is balanced. The second step generates a classification model for 
predicting the classes in the target dataset. The classification model uses as input a domain independent latent space learned from the source and augmented target datasets using our adversarial DA approach.

\subsection{Data Augmentation using the Autoencoder}\label{autoencoder}
An autoencoder is a neural network that is trained to reconstruct its input. It has two components: an encoder $enc$ that produces a compressed latent space $ h = enc(\bf{x})$ for input $\bf{x}$, and a decoder $dec$ that produces a reconstruction $\hat{\bf{x}} = dec(h)$. The objective is to minimize the following reconstruction error: 
\begin{align}\label{MSE}
\mathcal{L}(\mathbf{x}, dec(enc(\mathbf{x}))) = \Arrowvert \mathbf{x}-\hat{\mathbf{x}} \Arrowvert_2^2 = \Arrowvert \mathbf{x} - dec(enc(\mathbf{x}))\Arrowvert_2^2
\end{align}

 Autoencoders can be trained with minibatch gradient descent. At each batch, we feed the autoencoder with some data and backpropagate the error through the layers to adjust the weights of the networks. Although autoencoders can extract useful information from the data, they can cheat by copying the input to the output without learning useful properties of the 
data~\cite{goodfellow2016deep}. One way to prevent copying task is to use \textit{undercomplete autoencoders}, where the latent space has a smaller dimension than the input. With a reduced dimension, the autoencoders are forced to learn the most important attributes of the data. 

To generate synthetic data using an undercomplete autoencoder, we first train the autoencoder using the loss function in \eqref{MSE} with target data.    Then, the algorithm takes an original data as the input to our trained autoencoder and maps the original data to the compresses representation. Instead of passing the generated representation to the decoder, we 
add random noise drawn from a standard Gaussian distribution to the representation and pass it through the decoder to generate new synthetic data. The new data is labeled as the same class as the original data. To obtain a balanced training dataset, one can iterate the algorithm many times for the classes that have less samples.  Finally, we obtain an augmented target data by combing the origin data and synthetic data generated from above procedure. The augmented data is used by the DA algorithms described below.

\remove{
\begin{figure}[t]
\centerline{\includegraphics[width=\columnwidth]{TII-Articles-LaTeX-template/figures/autoencoder.drawio.pdf}}
\caption{Generation of synthetic data using an autoencoder.}
\label{fig1}
\end{figure}

\subsubsection{Autoencoder architecture} 
Our autoencoder consists of a encoder ($\mathit{Enc}$) and decoder ($\mathit{Dec}$) that are both CNNs (see Fig.~\ref{fig1}). The encoder has a convolution block with a \textsc{conv} layer, a \textsc{relu} activation layer and a \textsc{maxpooling} layer. The \textsc{conv} layer has $64$ $3 \times 3$ filters. To prevent the encoder copying the input to the output, the latent space has a smaller dimension than the input after the \textsc{maxpooling} layer. The latent space will be fed to the decoder that has a transposed convolution block and a convolution block. The transposed convolution block performs the transformation in the opposite direction of a normal convolution block with a \textsc{convt} layer and an \textsc{upsampling} layer. The last convolution block is needed to change the  dimension of the intermediate output and finally reconstruct the input image. To do so, we use a \textsc{conv} with $3$ filters which aligns with the input shape. We use a sigmoid layer as the final layer to reconstruct the input image. 
Finally, we train the autoencoder based on the following equation using only the target training dataset.  
\begin{align}\label{MSE}
L(x, dec(enc(x))) = \Arrowvert x-\hat{x} \Arrowvert^2 = \Arrowvert x - dec((enc(x))\Arrowvert^2
\end{align}

}

\subsection{Adversarial Domain Adaptation}\label{adv DA}


\subsubsection{Networks, inputs and outputs} 
Our architecture consists of five neural networks (see Fig.~\ref{fig1}): 1) $G_S$ is the private generator for the source; 2) $G_T$ is the private generator for the target; 3) $G$ is the shared generator; 4) $D$ is the discriminator 5) $C$ is the classifier. Note that for simplicity, the name of the neural networks includes the network architecture and all its weights. 

The source data is given by $(\mathbf{x^s}, \mathbf{y^s}, \mathbf{d^s})$ where $\mathbf{x^s}$ stands for the source data samples, $\mathbf{y^s}$ is the label  and $\mathbf{d^s}$ is the domain identity for source (e.g. $d_i^s = 0$ for any source sample $x_i^s$). Similarly, the target data is given by $(\mathbf{x^t}, \mathbf{y^t}, \mathbf{d^t})$ where $\mathbf{x^t}$ stands for the target data samples, $\mathbf{y^t}$ is the label and $\mathbf{d^t}$ is the domain identity for target (e.g. $d_i^t=1$ for any target sample $x_i^t$). Further more, let $N_s$ represent the number of samples from source domain and $N_t$ represent the total number of samples from target domain. We assume $N_s \gg N_t$. 

$\mathbf{x^s}$ and $\mathbf{x^t}$ are the inputs to private generators $G_S$ and $G_T$ respectively. $G_S$ and $G_T$ are separate networks so  the input $\mathbf{x^s}$ and $\mathbf{x^t}$ can have different dimensions. Shared generator $G$ learns the domain-independent representation ($\mathit{DI}$) from the outputs of $G_S$ and  $G_T$. Hence, the private generators must have the same shape of output vectors. The $DI$ output with corresponding networks is: 
\begin{equation}\label{eq2}
    \mathit{DI^s = G(G_S(\mathbf{x^s})) },\ \mathit{DI^t = G(G_T(\mathbf{x^t})) }
\end{equation}

The $DI$ is then used as the input for networks $D$ and $C$. The outputs of the two networks are $\mathbf{\hat{d}}$ from the discriminator $D$ and $\mathbf{\hat{y}}$ from the classifier $C$. 
\begin{equation}\label{eq3}
    \mathbf{\hat{d}} = D(DI),\  \mathbf{\hat{y}} = C(DI)
\end{equation}

\begin{figure}[t]
\centering
\includegraphics[width=\linewidth]{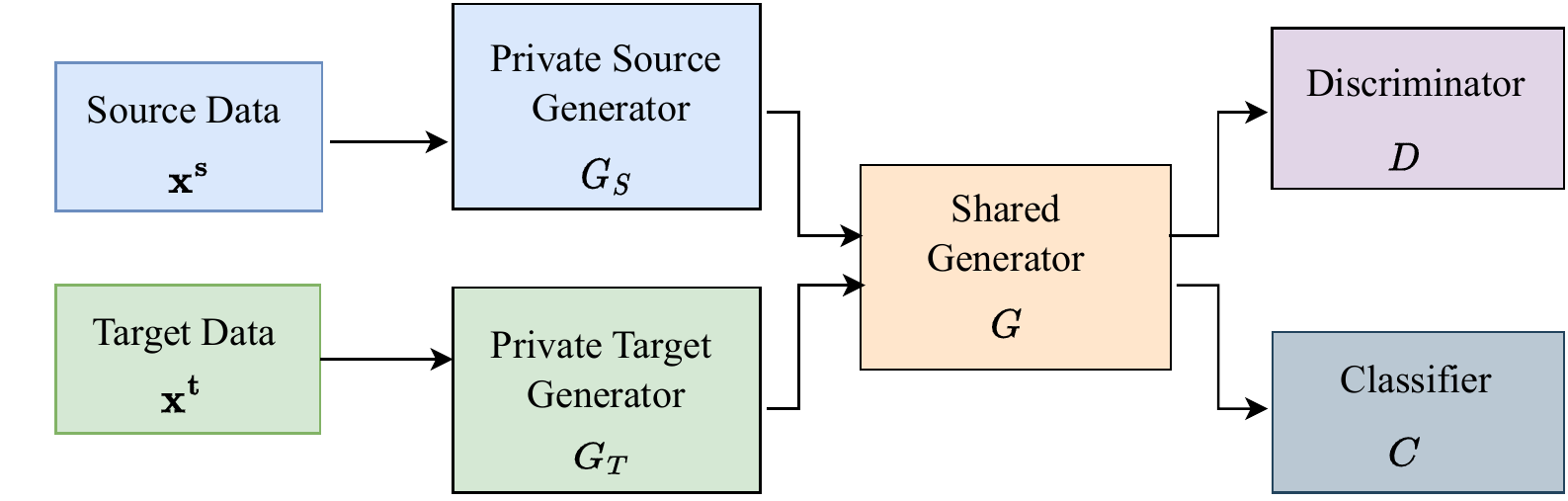}
\caption{Illustration of the proposed DA algorithm}
\label{fig1}
\end{figure}

\subsubsection{Loss functions and training} 
The classification loss is defined by the following expression, which measures the error of label prediction in both domains (we consider sufficient labeled data in the source and limited labeled data in the target). 
\begin{equation}
\mathcal{L}_c= 
 -  \sum_{i=1}^{N_s} {y}_i^s \cdot log \hat{y}_i^s - \lambda\sum_{i=1}^{N_t} {y}_i^t \cdot log \hat{y}_i^t 
\end{equation}%
where $y_i^s$ and $y_i^t$ are the one-hot encoding of the label for source input $x_i^s$ and target input $x_i^t$ respectively.  $\hat{y}_i^s$ and $\hat{y}_i^t$ are the softmax outputs of $C$. We use $\lambda$ as the penalty coefficient for the loss value obtained from
target data points. A good classifier should predict correct labels for source and target data points; therefore we minimize $\mathcal{L}_c$. 

The discriminator loss trains the discriminator to distinguish whether the $DI$ is generated from the source or the target data. $d_i$ is the domain identity for data $x_i$ ($d_i \in \{0,1\}$) and $\hat{d}_i$ is the output of discriminator $D$. The objective of the discriminator is to reduce the domain classification error; therefore we minimize $\mathcal{L}_d$. 
\begin{equation}\label{5}
\mathcal{L}_d= 
 -  \sum_{i=1}^{N_s + N_t} \left\{ {d}_ilog \hat{d}_i + (1-{d}_i)log (1-\hat{{d}}_i) \right\} 
\end{equation}%

The generator loss is the loss in \eqref{5}  with inverted domain truth labels. By minimizing $\mathcal{L}_g$, the generators
are trained in an adversarial manner to maximize the loss of
the discriminator.
\begin{equation}
\mathcal{L}_g= 
 -  \sum_{i=1}^{N_s + N_t} \left\{ (1-{d}_i)log \hat{d}_i + d_ilog (1-\hat{{d}}_i) \right\} 
\end{equation}%

The crux of successful DA is to learn features across domains that are predictive and domain invariant. A rich domain-independent representation should contain sufficient information needed for effective classification, no matter which domain the input data is from. To achieve domain invariance, we adversarially train a discriminator and several generators. To drive predictive information in the $DI$, we also train the generator to minimize the classification loss. In the following paragraph, we detail the training algorithm.

Training $G_S$, $G_T$, $G$ consists of optimizing $\mathcal{L}_g$ and $\mathcal{L}_c$, since we want to minimize the domain classification accuracy and maximize label classification accuracy. The discriminator is trained with $\mathcal{L}_d$ to maximize domain classification accuracy. We use $\mathcal{L}_c$ to train the classifier. Our learning algorithm follows the mini-batch gradient descent procedure. Such a procedure consists of  selecting an equal number of source and target samples, computing the outputs and loss functions, and adjusting the weight to the opposite direction to the gradient vector. This same process is iterated until the loss functions stop deceasing. More specifically,  the following steps are executed after creating the mini-batches of fixed size. The generators updates their weight to minimize the generator loss and classification loss as in Equations~\ref{12} - \ref{14}. The classifier updates its weight to minimize classification loss based on Equation~\ref{16}. The discriminator weights remain frozen during this step. Then, the discriminator updates its weight to minimize discriminator loss according to Equation~\ref{15}.
\begin{align}
     \label{12}
     \Delta_{G_S} & = - \mu \left( \beta\frac{\partial \mathcal{L}_g}{\partial G_S} + \gamma\frac{\partial \mathcal{L}_c}{\partial G_S}\right)\\ 
     \label{13}
     \Delta_{G_T} & = - \mu \left( \beta\frac{\partial \mathcal{L}_g}{\partial G_T} + \gamma\frac{\partial \mathcal{L}_c}{\partial G_T}\right) \\ 
     \label{14}
     \Delta_{G} & = - \mu \left( \beta\frac{\partial \mathcal{L}_g}{\partial G} + \gamma\frac{\partial \mathcal{L}_c}{\partial G}\right) \\ 
     \label{15}
     \Delta_{D} & =  - \mu\frac{\partial \mathcal{L}_d}{\partial D}\\
     \label{16}
     \Delta_{C} & =   - \mu\frac{\partial \mathcal{L}_c}{\partial C}
\end{align}
where $\mu$ is the learning rate.  The hyperparameters $\beta, \gamma$ are the relative weights of the loss functions.

\section{Experiment Details} 

We apply our pipeline to wafer defect prediction. 
Wafer test is an important step in semiconductor manufacturing, which involves evaluating the dies in a wafer and filtering out the defective ones. Past work has used machine learning (ML) approaches to expedite the prediction process~\cite{kang2015using}. However, as our experiments show, real-life wafer data suffers from low quality such as lack of labels and imbalanced class distribution, which would make most ML methods not suitable. In the experiments,  we also compare our approach against existing algorithms including fine-tuning based methods~\cite{b3} and DL based methods.

\subsection{Wafer Datasets}
\subsubsection{Source dataset} We use MixedWM38\footnote{https://github.com/Junliangwangdhu/WaferMap} dataset as the source dataset. MixedWM38 has 1 normal pattern, 8 single defect patterns, and 29 mixed defect patterns, with approximately 1000 samples for each category.  These wafer maps were obtained in a wafer manufacturing plant. Each wafer map is of size 52 $\times$ 52. 
 MixedWM38 does not have missing labels and the data size is consistent. The training data set is also balanced.

\subsubsection{Target dataset}
We use the WM-811K dataset\footnote{https://www.kaggle.com/datasets/qingyi/wm811k-wafer-map} 
as target. It comprises 811457 wafer maps collected from 46293 lots in real-world fabrication. It includes 8 single defect patterns and 1 normal classes which also appear in MixedWM38. However, the WM-811K dataset exhibits three common problems that can be found in manufacturing datasets. The first problem is that the dataset has a large amount of unlabeled samples. Approximately only 20\% of the wafer maps  are labeled from one of the nine types and can be used for learning. Second, the labeled wafer maps have different sizes. 
Finally, the dataset is highly imbalanced.

To solve the first two problems, we simply drop the wafer maps with no labels and select wafer maps of size $26 \times 26 $ in the remaining data. We choose this size because this is the only size group that has some data in each class.  There are $14,366$ wafer maps left in total: $90$ in center, $1$ in donut, $296$ in edge-loc, $31$ in edge-ring, $297$ in local, $16$ in near-full, $74$ in random, $72$ in scratch and 13489 in normal. For each class other than donut, we randomly select $60\%$ wafer maps in the training set and include the rest in the testing set. The two sets are disjoint except for sharing the same data in the donut class because there is only one sample available and we still want to include such pattern in classification. 

To address the third issue, that is, the imbalanced training data, we use the autoencoder-based data augmentation method introduced in Section~\ref{autoencoder}. The encoder has one $64$ $3 \times 3$ \textsc{conv} layer, one \textsc{relu} activation layer and one \textsc{maxpooling} layer. The decoder has one $64$ $3 \times 3$ \textsc{convt} layer, one \textsc{upsampling} layer, one $3$ $3 \times 3$ \textsc{conv} layer and a \textsc{sigmoid} output layer.  We generated $2000$ synthetic wafer maps for each defect class in the training set. We skipped normal class because the training set already has many data in that class. Note that the data augmentation only uses the WM-811K training data without seeing the WM-811K testing data. 

\subsection {Description of Experiments}\label{models}
We compare our pipeline under different settings and with different approaches. The methods we consider are: 
\remove{

The performance metrics we use are appropriate for evaluating a model on an imbalanced data set where the dataset has a lots of negative classes (in our example the majority of wafer maps are in none). We thus use the following metrics:
\begin{itemize}
\item {\bf Balanced accuracy.} The conventional accuracy is defined as the total number of correct predictions divided by the total number of predictions made for a dataset. When working with an imbalanced data set, accuracy does not tell the full story. A model can achieve a high accuracy if it just predicts every sample as the majority class.  On the other hand, the balanced accuracy~\cite{balanced_accuracy} is designed to work well with imbalanced data. It is defined as \textit{the average of recall obtained on each class}, which is calculated as the sum of true positives divided by the sum of true positives and false negatives. Since the data set has a very disproportionately high number of negative cases, the model may detect a larger number of positive cases as negative.

\item {\bf Precision.} It is calculated as the sum of true positives across all classes divided by the sum of true positives and false positives across all classes. If there is a high number of false positives, the precision will become low.

\end{itemize}
}

\subsubsection{\textbf{Adversarial DA + augmented target data}} We use the MixedWM38 training data as the source training data and the augmented WM-811K training data as the target training data. These data is used as input to our adversarial DA networks which are subsequently trained based on the process described in section \ref{adv DA}. This is our proposed approach.

In the architecture used in the experiments, $G_S/G_T$ has two convolutional layers: $8\ 5\times5$ filters (\textsc{conv1}), $16\ 5\times5$ filters (\textsc{conv2}), two maxpooling layers of size $2\times2$ after \textsc{conv1} and \textsc{conv2} respectively, and one fully connected layer with $2028$ neurons.  $G$ has the same configuration as $G_S$ and $G_T$ but the last fully connected layer has only $1024$ neurons and we add a reshape layer of $(26,26,3)$ at the beginning of the network. The configuration of $D$ is similar to $G$ but it has a softmax output layer for domain prediction. The classifer has two fully connected layers with $1024$ and $512$ neurons respectively and a \textsc{softmax} output layer for class prediction.

\subsubsection{\textbf{Adversarial DA + imbalanced target data}}
We still use the adversarial DA networks but replace the target training data with the imbalanced WM-811K training data without augmentation. Comparing this approach with approach 1) determines whether our data augmentation step improves the performance of adversarial DA.

\subsubsection{\textbf{Fine-tuning + augmented target data}}We use the fine-tuning approach by Shao et al.~\cite{b3} for transferring knowledge learned 
from general images to identify  machine faults from images of induction motors, gearboxes, and bearings. They use a pre-trained VGG $16$ model trained on ImageNet ~\cite{imagenet}. VGG $16$ has $5$ convolution blocks and a fully connected block. They freeze the first three convolution blocks and retrain the last two convolution blocks and the fully connected block using the machine fault diagnosis dataset.  Cross entropy loss serves to evaluate errors between true labels and predicted probabilities. We implement their approach but replace the machine fault dataset with the augmented WM–811K training dataset. The output layer of the pre-trained VGG $16$ model is replaced with a new layer with $9$ neurons corresponding to $9$ classes.

\subsubsection{\textbf{Fine-tuning + imbalanced target dataset}}
This approach is the same as the previous one 
but uses the imbalanced WM–811K training dataset.  Comparing this approach with the previous one  determines whether our data augmentation step benefits the fine-tuning approach.

\subsubsection{\textbf{Vanilla classifier + augmented target dataset}}
We train a deep neural network that serves as a classifier to detect defects in the wafer maps.  The network is trained with the augmented WM-811K training data using the cross entropy loss. The classifier has an architecture compatible with the prediction pipeline used in our DA method so the comparison numbers are fair and meaningful.  The classifier has $3$ convolution blocks and two fully connected blocks. Each convolution block has a \textsc{conv} layer and a \textsc{relu} layer. The convolutional layers have an increasing output filters of $\{16, 64, 128\}$ respectively. Each fully connected block has a \textsc{fc} layer and a \textsc{relu} activation layer. The first \textsc {fc} layer has $512$ neurons and the second \textsc{fc} layer has $128$ neurons. The output layer has $9$ neurons, followed by a \textsc{softmax} layer for predicting probabilities for each class. 

\subsubsection{\textbf{Vanilla classifier + imbalanced target dataset}}
We train the same deep neural network of case 5) using the imbalanced WM– 811K training data.

\section{Results and Analysis}

\begin{figure}[t]
\centerline{\includegraphics[width=\columnwidth]{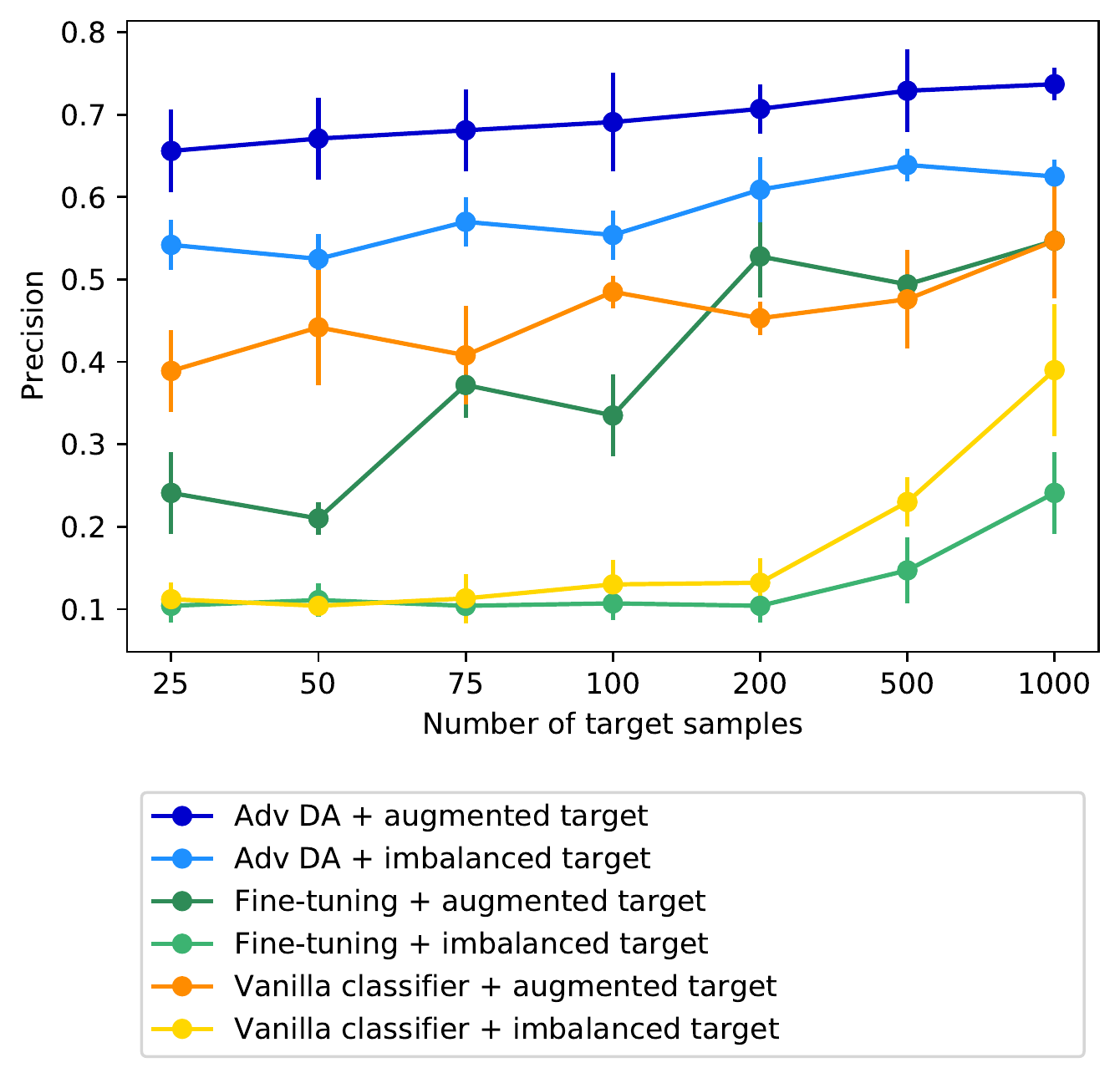}}
\caption{Classification precision score achieved using augmented target data and imbalanced target data comparing six approaches: a vanilla deep CNN trained with the augmented/imbalanced target samples; a pretrained VGG 16 model fine-tuned with the augmented/imbalanced target data; our adversarial DA architecture trained with augmented/imbalanced target data.}
\label{fig2}
\vspace{-15pt}
\end{figure}

We use the TensorFlow~\cite{abadi2016tensorflow} and Keras~\cite{KerasDocumentation} libraries for training our adversarial DA and the other classification models. 
For our adversarial DA, we train for $20000$ iterations with a batch size of $32$. We use the Adam optimizer with a starting learning rate of $2e-4$ and set the hyperparamters $\lambda = 0.1, \beta = 1, \gamma = 1$ (the hyperparameters were not tuned using  validation samples). For the fine-tuning approach, the VGG $16$ pre-trained model implemented by Keras requires the input to have exactly $3$ input channels, and width and height not to be smaller than $32$. The input size of the target training data is $26\times26\times3$, thus is an invalid value. We use the Python resize function to change the target input to $32\times32\times3$. For the fine-tuning approach and vanilla classifier method, we use $60$ epochs for training with a batch size of $32$ and the Adam optimizer with a learning rate of $2e-4$. We apply an early stopping technique in training that saves the best weight by comparing the performance in each epoch.

\begin{table*}[t]
\caption{Balanced Classification Accuracy/Averaged Recall Comparison on the WM-811K Testing Data}
\label{table1}
\setlength{\tabcolsep}{3pt}
\centering
\begin{tabular}{lrrrrrrr}
\toprule
\multicolumn{1}{c}{\multirow{2}{*}{Models in \ref{models}}} &\multicolumn{7}{c}{Number of samples in target dataset used for training} \\ \cline{2-8}

\multicolumn{1}{c}{}& \multicolumn{1}{c}{25}  & \multicolumn{1}{c}{50}  & \multicolumn{1}{c}{75}  & \multicolumn{1}{c}{100}  & \multicolumn{1}{c}{200}  & \multicolumn{1}{c}{500}  & \multicolumn{1}{c}{1000}    \\
\midrule
Adversarial DA  + augmented & 70.3\% $\pm$ 0.7\%
                            &71.7\% $\pm$ 4.5\%
                            &71.2\% $\pm$ 4.5\%
                            &72.5\% $\pm$ 4.3\%
                            &72.5\% $\pm$ 2.6\%
                            &72.3\% $\pm$ 4.2\%
                            &73.7\% $\pm$ 2.5\%\\
\midrule
Adversarial DA  + imbalanced &54.7\% $\pm$ 4.6\%
                            &57.9\% $\pm$ 5.6\%
                            &56.3\% $\pm$ 4.5\%
                            &65.2\% $\pm$ 4.2\%
                            &67.2\% $\pm$ 1.4\%
                            &65.9\% $\pm$ 2.7\%
                            &65.8\% $\pm$ 0.1\%\\
\midrule
Fine-tuning   + augmented & 28.9\% $\pm$ 8.3\%
                          & 28.3\% $\pm$ 7.6\%
                          & 45.5\% $\pm$ 5.8\%
                          & 49.1\% $\pm$ 7.6\%
                          & 56.1\% $\pm$ 2.4\%
                          & 65.3\% $\pm$ 3.1\%
                          & 67.4\% $\pm$ 1.2\% \\
\midrule
Fine-tuning  + imbalanced & 11.1\% $\pm$ 0\%
                            &  13.3\% $\pm$ 6.1\%
                            & 11.1\% $\pm$ 0.0\%
                            & 13.2\% $\pm$ 5.9\%
                            & 11.1\% $\pm$ 0.0\%
                            & 15.2\% $\pm$ 4.6\%
                            & 24.8\% $\pm$ 7.1\%\\
\midrule
Vanilla classifier  + augmented & 48.7\% $\pm$ 3.8\%
                                & 53.1\% $\pm$ 7.3\%
                                & 52.1\% $\pm$ 1.7\%
                                & 63.7\% $\pm$ 7.3\%
                                & 63.6\% $\pm$ 7.4\%
                                & 65.3\% $\pm$ 7.0\%
                                & 65.3\% $\pm$ 7.7\%\\
\midrule
Vanilla classifier + imbalanced & 11.4\% $\pm$ 1.0\%
                                & 11.1\% $\pm$ 0.0\%
                                & 11.4\% $\pm$ 1.5\%
                                & 11.7\% $\pm$ 1.9\%
                                & 13.2\% $\pm$ 7.5\%
                                & 27.3\% $\pm$ 8.1\%
                                & 35.5\% $\pm$ 6.5\%\\
\bottomrule

\end{tabular}
\label{tab1}
\end{table*}

\begin{table}[t]
\caption{Training and testing time comparison for three methods. The results are obtained with 1000 target data sampled from the augmented target training data. We can train the proposed DA model offline. The prediction time is as low as a vanilla classifier. }
\label{time}
\centering
\begin{tabular}{@{}lrr@{}}
\toprule
\multicolumn{1}{c}{Models} & Training time (s) & Testing time (s) \\ \midrule
Adversarial DA             & 3211.70      & 0.47       \\ \midrule
Fine-tuning                & 31.71       & 0.87       \\ \midrule
Vanilla classifier         & 86.33       & 0.45       \\ \bottomrule
\end{tabular}
\end{table}

For this evaluation, the source training dataset has $5,294$ wafer maps from $9$ categories with an even distribution. All these experiments are performed for target training datasets containing only $25, 50, 75, 100, 200, 500$ and $1000$ randomly selected samples. The goal of these experiments is to show the impact of target training data size on the performance of different models. Balanced classification accuracy and precision are calculated on the target testing data and the 95\% confidence intervals  are shown in Tables~\ref{table1} and Fig.~\ref{fig2} with comparisons to methods presented in Section~\ref{models}. Those intervals are obtained from $5$ repeated experiments. Table \ref{time} shows the training and testing time of different approaches. The performance metrics are appropriate for evaluating a model on an imbalanced dataset. The balanced accuracy is designed to work well with imbalanced data. It is defined as the average of recall obtained on each class, which is calculated as the sum of true positives divided by the sum of true positives and false negatives.  On the other hand, precision is calculated as the sum of true positives across all classes divided by the sum of true positives and false positives across all classes. If there is a high number of false positives, we will have a low precision score.

For $25-1000$ samples, we observe that our adversarial DA with augmented target approach outperforms the fine-tuning methods and the deep CNN methods in terms of balanced accuracy and precision. The performance of our method and deep neural network will improve further if we use more sophisticated models such as ResNet~\cite{he2016deep}. Nevertheless, with comparable architectures, the inferior performance of the vanilla classifier approach when trained with very little data shows the limit of the DL approach: a large number of training data is required to learn the input-out function of the model. Failure to do so can lead to the well known over-fitting issue where the model memorizes the training data; hence it does not generalize well on the new testing data. As a TL approach, fine-tuning makes use of a pre-trained model so the network obtains some sensible weights that can be transferred to the target task. However, it does not directly address the problem of scarcity of  training data.  We postulate that the poor performance of the fine-tuning approach in our experiments may stem from the large difference between ImageNet and wafer maps which requires a reasonable amount of target data to successfully update the weights of the pre-trained model. On the other hand, our adversarial DA approach achieves the best results because it alleviates the problem of the small amount of target training data by using domain invariant features emerging in the course of the optimization. Given sufficient balanced source data, those features can be learned even with very limited target data by our adversarial learning framework.

Across all three methods, training with augmented target data gives them a significant performance boost, showing the effectiveness of our data augmentation technique in dealing with highly imbalanced data. For example, if we use the augmented target for adversarial DA approach, the balanced accuracy  increases by $5\% - 16\%$ for $25-1000$ samples and the precision increases by $6\%-15\%$.  The observation is even more evident for the fine-tuning and vanilla classifier methods. With the imbalanced target, the fine-tuning method fails to learn anything useful. On the other hand,  the evidence that our DA approach outperforms other methods even in the case of using 1000 augmented target samples for training confirms that our DA approach generalizes better on the target test data (real data).

\section{Using DA for non-classification tasks}

Our approach for addressing the training data scarcity issue
can be extended to non-classification tasks. In this section, we briefly discuss recent approaches for handling the domain shift and accomplish effective knowledge transfer in the area of optimization, reinforcement learning and robotic learning. 

In the area of transfer optimization (TO)~\cite{jiang2017transfer,bali2019multifactorial}, solutions from various source optimization problems are utilized to solve a target optimization task. The approach by Jiang et al.~\cite{jiang2017transfer} integrates a DA method to a classical evolutionary optimization algorithm to improve the search efficiency for dynamic optimization problems.
Approaches have also been proposed to model the function to be optimized, that is, the objective function via an artificial neural network (ANN)~\cite{Villarubia}. Such approximation is useful, for example, to reduce computational costs.
However, these approaches require training the ANN using input-output pairs generated from a known function. If the underlying target function is unknown and there are only limited measurements available, our DA can be used in the sense that the input-output pairs from a known function may be used as the source domain to guide the learning of the target domain, i.e., the very limited measurement samples governed by an unknown function.

In reinforcement learning (RL), one of the open problems is that the input data distribution may change over time, so the learnt polices may not work well with the new input data. 
Recent approaches have been proposed that apply DA to enable RL agents to be effective even if the input distribution changes over time~\cite{carr2018domain,higgins2017darla}. In this scenario, the source domain is a particular input distribution with a specified reward structure. The target domain has a modified input distribution but the reward structure is the same. Domain shift is also a major challenge in learning-based robotic perception and control. 

Robots trained using simulated data often fail in  real world environments due to the sim-to-real gap. The approach by Tzeng et al.~\cite{tzeng2020adapting} shows successful adaptation of pose estimation from synthetic images to real images by using domain confusion loss (similar to $\mathcal{L}_g$) and pairwise loss together.

\section{Limitations}
While we use available data from a single source domain to improve the generalization on a related target task, one may find data from many related domains useful. For example, one can have several labeled manufacturing datasets collected over time or from different parties to use as the source domain. Our current approach does not directly support multi-source domain adaptation. To use our approach in a multi-source setting, one needs to combine all the source data as one source domain or train with each source domain individually and pick the one with the best performance. A better approach is to treat each source domain as individual domain and learn the shared information across different domains. Work along this direction~\cite{zhao2018multiple, peng2019moment} has shown better generalization performance on the target than the single-source approach. 

Another limitation of our approach is that it requires at least some labeled data from each class in the target domain. The reason is that the autoencoder-based data augmentation procedure requires the original target to have labeled data from each category to construct a balanced target dataset. Our adversarial DA approach can be used alone in the unsupervised setting where the target data has no label, conditioning that the target data is balanced.

\section{Conclusion}

This paper proposes a novel adversarial DA approach supporting heterogeneous adaptation where the source domain has different features than the target domain. DA is accomplished by  training two private generators and a shared generator that extract features which are predictive of the target labels, but uninformative about the domain.   While the DA approach aims to tackle the problem of scarcity of target training data, it does not work well if the target data is imbalanced.  Collecting balanced data is challenging with many manufacturers facing the reality of low quality data. To address this issue,  we further propose an pipeline using an autoencoder-based technique for augmenting minority classes in the training data, followed by our DA approach. The experimental evaluation of our pipeline on wafer defect datasets demonstrates its superior performance compared with other baseline approaches. 

\noindent
{\bf Acknowledgments.} The work reported in this paper has been funded by NSF under grant CMMI 2134667 ``Privacy-Preserving Tiny Machine Learning Edge Analytics to Enable AI-Commons for Secure Manufacturing.''

\balance
\bibliographystyle{plain}
\bibliography{main}

\end{document}